\documentclass[11pt]{article}

\usepackage[preprint]{acl}

\usepackage{times}
\usepackage{latexsym}
 
\usepackage[T1]{fontenc}
\usepackage[utf8]{inputenc}

\usepackage{microtype}

\usepackage{inconsolata}

\usepackage{graphicx}

\usepackage{amsmath}
\usepackage{amssymb}
\usepackage{pifont}
\newcommand{\xmark}{\ding{55}}
\usepackage{multirow}
\usepackage{listings}
\usepackage[most]{tcolorbox}
\usepackage{upquote}
\usepackage{booktabs} 

\definecolor{PromptBack}{HTML}{F7F8FA}   
\definecolor{PromptFrame}{HTML}{4B5563}  
\definecolor{PromptTitle}{HTML}{111827}

\lstdefinestyle{promptlisting}{
  breaklines=true,
  breakatwhitespace=false,
  postbreak=\mbox{\textcolor{PromptFrame}{\(\hookrightarrow\)}\space},
  basicstyle=\ttfamily\scriptsize, 
  columns=fullflexible,
  keepspaces=true,
  tabsize=2,
  showstringspaces=false,
  xleftmargin=0pt,
  xrightmargin=0pt
}
\newtcolorbox{promptbox}[1][]{
  enhanced, breakable,
  colback=PromptBack,
  colframe=PromptFrame,
  coltitle=PromptTitle,
  fonttitle=\bfseries\footnotesize,
  fontupper=\footnotesize,             
  title=#1,
  boxrule=0.6pt,                       
  arc=1.5pt,
  left=4pt,right=4pt,top=4pt,bottom=4pt,  
  before upper={\setlength{\parindent}{0pt}\setlength{\parskip}{2pt}}
}

\title{LayoutCoT: Unleashing the Deep Reasoning Potential of Large Language Models for Layout Generation}

\author{
  Hengyu Shi\thanks{\ \ Both authors contributed equally to this research. Names are ordered alphabetically.},
  Junhao Su\footnotemark[1],
  Tianyang Han,
  Junfeng Luo,
  Jialin Gao\thanks{Corresponding author.}\\
  Meituan\\
  \texttt{\{shihengyu02,sujunhao02,hantianyang,luojunfeng,gaojialin04\}@meituan.com}
}

\AtBeginDocument{
    \setlength{\abovedisplayskip}{3pt plus 1pt minus 1pt}
    \setlength{\belowdisplayskip}{3pt plus 1pt minus 1pt}
    \setlength{\abovedisplayshortskip}{0pt plus 1pt}
    \setlength{\belowdisplayshortskip}{2pt plus 1pt}
}

\begin{document}

\maketitle

\begin{abstract}
Conditional layout generation aims to automatically produce visually appealing and semantically coherent layouts under user-defined constraints. 
Generative models excel but require large task-specific datasets or costly fine-tuning, hampering scalability and real-world adoption. 
Training-free LLM methods, though cheaper, still suffer from limited reasoning depth and naive ranking strategies, leading to sub-optimal layouts.
To this end, we introduce LayoutCoT, the first framework that explicitly unlocks the latent reasoning abilities of off-the-shelf LLMs for layout generation by uniting Layout-aware Retrieval-Augmented Generation (RAG) with Chain-of-Thought (CoT) reasoning. 
LayoutCoT first linearizes layouts into a LLM-friendly serialized form. Then it employs a layout-aware RAG module to retrieve relevant exemplars and then prompt the LLM to draft a coarse layout with the user's constraints. Finally, we introduce a dedicated CoT module to iteratively refine this draft, reasoning step-by-step over spatial and semantic relations to produce layouts with markedly higher coherence and visual quality.
On five public benchmarks covering three conditional layout generation tasks, LayoutCoT establishes new state-of-the-art results without fine-tuning.
Remarkably, our CoT module enables a standard LLM to surpass specialized deep-reasoning models such as DeepSeek-R1, underscoring the promise of CoT-driven LLMs for scalable, training-free layout generation.
\end{abstract}

\begin{figure*}[t]
  \centering
  \includegraphics[width=0.85\textwidth]{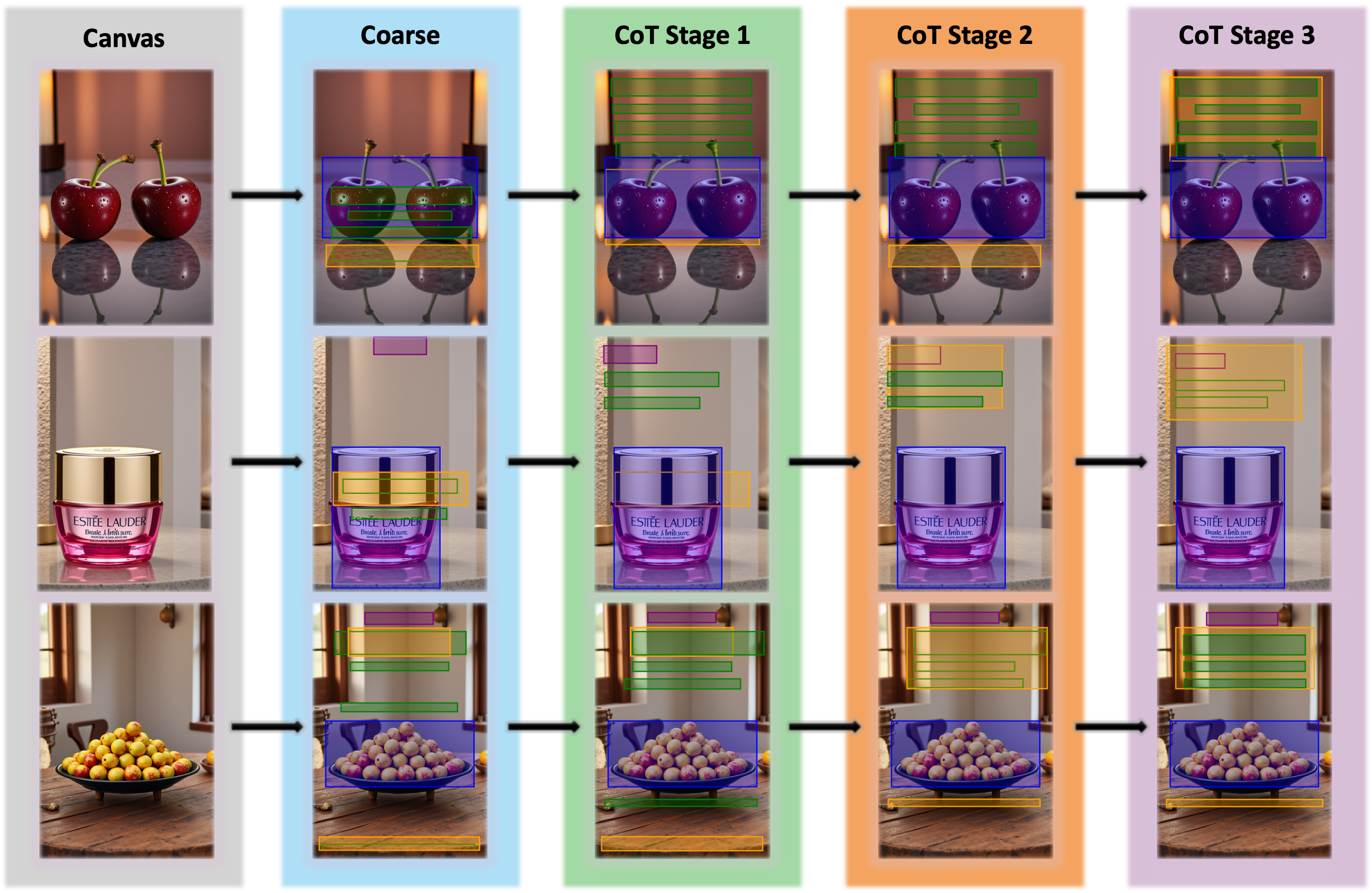}
  \caption{Qualitative evaluation of the multi-stage LayoutCoT framework. Each successive stage systematically refines the spatial arrangement of layout elements, resulting in designs with enhanced coherence and rationality.}
  \label{Fig.1}
\end{figure*}

\section{Introduction}
Layout is a fundamental element in graphic design, determined by the deliberate and structured arrangement of visual components. Automated layout generation has recently emerged as an important research topic, aiming to significantly alleviate the workload of designers while addressing diverse user requirements \cite{1,2,3,4,5,6}. Current automated layout generation tasks broadly fall into three categories: i) content-aware layouts, which integrate visual or textual content for contextually appropriate designs (e.g., aligning text boxes with salient image regions in posters); ii) constraint-explicit layouts, which strictly adhere to predefined constraints, commonly found in user interface (UI) design and document typesetting; iii) and text-to-layout tasks, which translate abstract textual instructions into spatially coherent layouts, crucial for cross-modal creative tools.

Early GAN and diffusion-based approaches~\cite{13,14} learned implicit design rules from extensive labeled datasets. Methods such as LayoutGAN \cite{7} and LayoutDiffusion \cite{8} have shown promising capabilities in generating plausible layouts, yet their reliance on substantial training data and domain-specific fine-tuning limits scalability and versatility. When faced with limited training data, purely generative methods \cite{2,9,10,11,12} often fail to effectively capture sparse data distributions, restricting generalization and necessitating separate models for distinct tasks. Furthermore, the demand for extensive labeled datasets makes training generative layout models costly and inefficient, highlighting the necessity for a more flexible and versatile layout generation paradigm.

With the advancement of large language models (LLMs) \cite{15,16,17,18,19}, layout generation methods are shifting towards a training-free paradigm. Recent research has shown that LLMs inherently possess a certain understanding of layout principles, such as alignment and spatial coherence \cite{16,20}.
While leveraging this capability, approaches such as LayoutPrompter \cite{12} utilize in-context learning combined with post-processing ranking mechanisms to guide LLMs in layout generation. However, this methodology demonstrates excessive focus on quantitative evaluation metrics, frequently resulting in structurally impractical bounding box configurations containing either disproportionately small elements or physically unfeasible arrangements. Moreover, when confronted with complex layout scenarios, the simplistic reasoning paradigms and elementary ranking systems inherent in such approaches tend to produce suboptimal compositions that violate both human visual perception principles and real-world application constraints.

To unleash the deep reasoning abilities of LLMs in layout generation, we must resolve two key issues: first, how to select suitable contexts as references for layout generation; and second, how to design a chain of thought that enables LLMs to engage in human-like deep reasoning during layout design. Therefore, we propose LayoutCoT, a novel approach that integrates a layout-aware RAG with Chain-of-Thought (CoT) in LLMs to address those above limitations. Specifically, LayoutCoT employs a layout-aware exemplar retrieval mechanism, dynamically selecting suitable reference contexts using layout similarity \cite{21}, which better captures structural and semantic similarities compared to previous methods like LayoutPrompter \cite{12}. 
We further introduce a meticulously designed Chain-of-Thought (CoT) reasoning framework that refines the layout through three sequential stages, each addressing a distinct aspect of the design process. In the first stage, the LLM is prompted to evaluate each element from the specified categories and determine their initial positions, emphasizing logical structure and visual appeal without accounting for interactions among element sizes. In the second stage, the LLM is guided to optimize the sizes and positions of these elements, effectively resolving overlaps or potential crowding issues. Finally, in the third stage, fine-grained mathematical adjustments are applied to specific layout dimensions to avoid unreasonable arrangements—such as the occlusion of salient elements—and to ensure both aesthetic quality and practical viability.

We conducted comprehensive experiments across five public datasets (PKU \cite{11}, CGL \cite{23}, RICO \cite{25}, Publaynet \cite{26}, and WebUI \cite{24}), spanning three distinct layout generation tasks. Our results demonstrate that LayoutCoT consistently outperforms existing state-of-the-art methods without requiring any training or fine-tuning. Notably, our CoT module enables standard LLMs, such as GPT-4, to surpass the performance of specialized deep-reasoning models like DeepSeek-R1 \cite{22}, underscoring the potential of our approach for practical layout generation tasks. The main contributions of our work are summarized as follows:
\begin{itemize}
    \item 
    We introduce LayoutCoT, a method that effectively harnesses the deep reasoning capabilities of general-purpose LLMs through Chain-of-Thought prompting, significantly improving the performance and practicality of training-free layout generation.
    \item 
    LayoutCoT requires no task-specific training or fine-tuning, efficiently producing high-quality, visually coherent layouts by decomposing complex generation tasks into clear reasoning steps.
    \item 
    Extensive empirical validation demonstrates LayoutCoT’s superior performance and versatility across various tasks and datasets, highlighting its ability to elevate standard LLMs above specialized deep-reasoning counterparts.
\end{itemize}

\begin{figure*}[!htbp]
    \centering
    \includegraphics[width=0.89\textwidth]{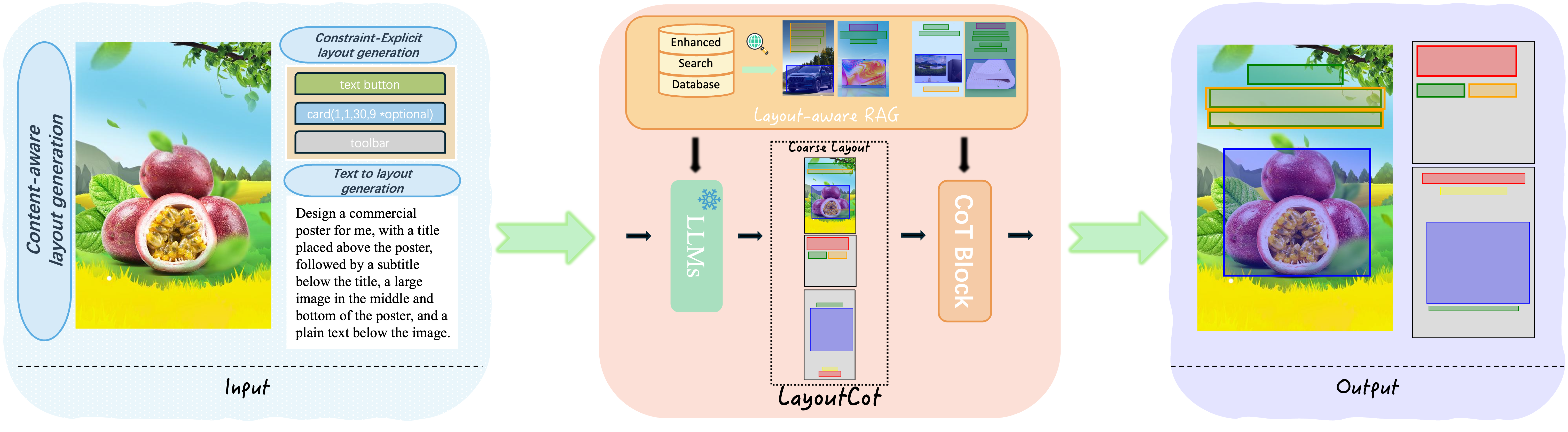}
    \caption{Overview of the LayoutCoT framework. Our training-free approach initially employs a Layout-aware RAG to prompt the LLM for a coarse layout prediction, establishing a logical and visually coherent arrangement of elements. This is subsequently refined via a multi-stage Chain-of-Thought (CoT) module, which iteratively enhances the layout by resolving spatial conflicts and fine-tuning dimensions. The proposed framework is versatile and applicable to a wide array of layout generation tasks.}
    \label{Fig.2}
\end{figure*}

\section{Related Work}
Layout generation has emerged as an active area of research in recent years. It can be broadly divided into three categories: content-aware layout generation \cite{11,27}, constraint-explicit layout generation \cite{3,4,5}, and text-to-layout generation \cite{28,29}. Content-aware layout generation aims to place layout elements on a given canvas in a visually pleasing manner while avoiding overlap with any background images on the canvas. Constraint-Explicit layout generation typically involves arranging predefined elements, or elements with specified dimensions, or even specified relationships among elements, on a blank interface in a reasonable configuration. Text-to-layout generation, meanwhile, focuses on creating layouts based on natural language descriptions, which is generally more challenging.

\subsection{Generative layout generation}
Early work in this field commonly utilized generative models for layout generation. Research based on Generative Adversarial Networks (GANs) \cite{6,11} and Variational Autoencoders (VAEs) \cite{5,30} pioneered the application of generative models to layout generation, achieving noticeable success in improving layout aesthetics through latent variable optimization. With the advent of the Transformer \cite{31} architecture, numerous studies \cite{1,4,28,29,32,49,50} emerged that adopted a sequence-to-sequence approach, further enhancing the quality of generated layouts. More recently, the proposal of diffusion \cite{14} models has led to additional methods \cite{9,10} that bolster generative capabilities in scenarios with explicit layout constraints, offering further advancements in both quality and overall visual appeal. However, the existing generative layout methods are not capable of handling all types of layout generation tasks using a single model. Moreover, these approaches typically rely heavily on extensive datasets and require substantial training or fine-tuning, making practical applications challenging.

\subsection{LLM-based Layout Generation}
Large language models have demonstrated exceptional few-shot performance \cite{15,17,33,34} across various natural language processing (NLP) tasks. Recent studies have shown that LLMs themselves can generate layouts in structured descriptive languages such as HTML \cite{12,35}. For instance, LayoutNUWA \cite{35} leverages Code Llama’s \cite{36} code completion for layout generation, while PosterLlama \cite{37} fine-tunes models such as MiniGPT-4 \cite{38} to further enhance code completion capabilities for layout generation tasks. However, these methods rely heavily on sophisticated fine-tuning strategies and are dependent on large amounts of data. In contrast, LayoutPrompter \cite{12} and LayoutGPT \cite{39} utilize GPT-3 \cite{15} via in-context learning to achieve training-free layout generation, revealing that LLMs inherently possess the capacity for layout generation. Although these methods do not require extensive, painstaking training and can generalize across various tasks, the quality of their generated layouts remains a concern.

\subsection{Retrieval-Augmented Generation and Chain of Thought}
Retrieval-Augmented Generation (RAG) \cite{40} is a cutting-edge technique that combines retrieval and generation to enhance the model's generative capabilities. By retrieving relevant examples or information \cite{41,42}, RAG introduces more comprehensive contextual data during the generation process, thereby improving the quality of the outputs. Content-aware graphic layout generation aims to automatically arrange visual elements based on given content, such as e-commerce product images. In this context, the Retrieval-Augmented Layout Transformer (RALF) \cite{43} leverages the retrieval of layout examples most similar to the input image to assist in layout design tasks. However, the RAG in RALF is constrained by its input paradigm, lacking plug-and-play characteristics and performing suboptimally across other tasks.
Chain-of-Thought (CoT) \cite{44,45,46} is a method that enhances model reasoning capabilities through the simulation of human thought processes. CoT achieves this by progressively solving problems through phased reasoning, offering more interpretable and coherent solutions to complex tasks. Although CoT has demonstrated powerful reasoning advantages across numerous domains, it has not yet been applied to layout design tasks. We attempt to integrate RAG with CoT for layout generation tasks, utilizing RAG to provide rich contextual information, while employing CoT for phased reasoning and optimization. This synergistic approach aims to enhance the quality and coherence of the generated results, combining the strengths of both methodologies effectively.

\begin{table}[!htbp]
\renewcommand{\arraystretch}{1.1}
\centering
\caption{Dataset statistics. Training Set refers to the candidate pool for RAG example selection.}
\small{
\resizebox{0.9\linewidth}{!}{
    \begin{tabular}{cccccc}
        \toprule
        Dataset &  Tasks &  Training Set &  Test Set &  Element Types \\
        \midrule
        PKU  & content-aware & 9,974 & 905 & 3 \\
        CGL  & content-aware & 38,510 & 1,647 & 5 \\
        RICO &  constraint-explicit & 31,694 & 3,729 & 25 \\
        PubLayNet  & constraint-explicit& 311,397 & 10,998 & 5 \\
        WebUI & text-to-layout & 3,835 & 487 & 10 \\
        \bottomrule
    \end{tabular}
}}
\label{Table.1}
\end{table}

\section{Methods}
Figure. \ref{Fig.2} illustrates the overall structure of LayoutCoT, which consists of three main components: the Layout-aware RAG, the layout coarse generator, and the layout CoT module. Specifically, given the labels and bounding boxes (bboxes) of an input example layout, we use the Layout-aware RAG to select the top-k relevant layout examples as prompting examples. Next, the layout coarse generator takes these prompting examples and the given constraints to produce an initial coarse layout. Finally, both the prompting examples and this coarse layout are fed into the layout CoT module for fine-grained layout adjustments.
Because the layout coarse generator lacks a detailed reasoning process, relying solely on the prompting examples and constraints may not fully satisfy the design requirements. Therefore, the layout CoT module breaks down the layout task into a multi-stage thought process; each stage refines the layout, ultimately yielding a more precise and aesthetically pleasing result.

\begin{figure*}[!t]
    \centering
    \includegraphics[width=0.8\linewidth]{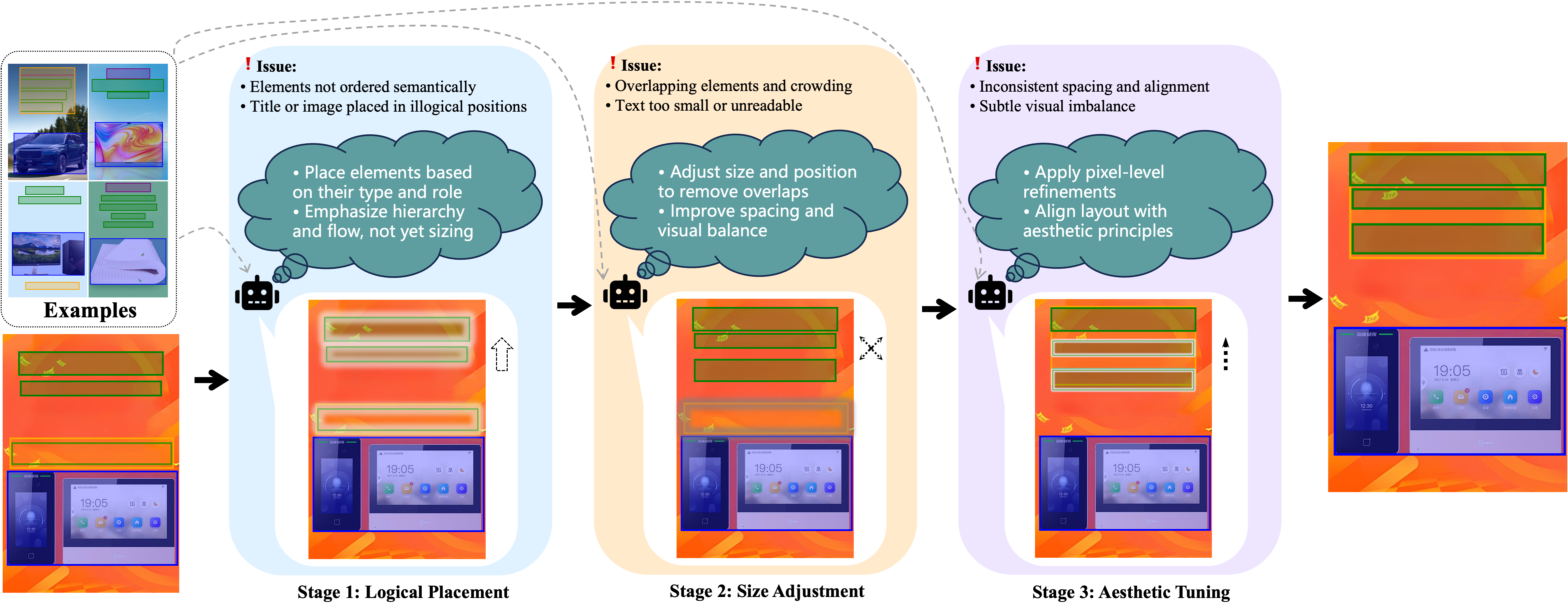}
    \caption{Details of the CoT Module. We illustrate the overall conversational logic of the multi-stage CoT. Depending on the type of task, there are slight variations in the details to adapt to specific requirements. Some of LayoutCoT prompts can be found in the supplementary material.}
    \label{Fig.3}
\end{figure*}

\subsection{Layout-aware RAG}
\label{sec3.1}
We employ a Retrieval-Augmented Generation (RAG) mechanism to dynamically select relevant layout examples as prompts, thereby enhancing the flexibility and coherence of generation. Specifically, let 
\(
\mathcal{L} = \bigl\{(b_i, c_i)\bigr\}_{i=1}^m
\) 
and 
\(
\hat{\mathcal{L}} = \bigl\{(\hat{b}_j, \hat{c}_j)\bigr\}_{j=1}^n
\) 
denote two layouts, where each element consists of a bounding box \(b_i\) (or \(\hat{b}_j\)) and an associated label \(c_i\) (or \(\hat{c}_j\)). We use a dissimilarity function from LTSim \cite{21} \(D(\mathcal{L}, \hat{\mathcal{L}})\) based on a \emph{soft matching} scheme:
\begin{equation}
D(\mathcal{L}, \hat{\mathcal{L}})
=
\min_{\Gamma \ge 0}
\sum_{(i,j)\in \Omega}
\Gamma_{i,j}\,
\mu\bigl(b_i, c_i, \hat{b}_j, \hat{c}_j\bigr)
\end{equation}
where \(\Gamma_{i,j}\) represents a fractional assignment aligning element \((b_i, c_i)\) with \((\hat{b}_j, \hat{c}_j)\), and \(\mu(\cdot)\) is a cost function that captures positional and label discrepancies. By solving this optimization, we obtain the minimal “transport cost” between the two layouts.

Next, we transform this cost-based measure into a layout-level similarity via an exponential mapping:
\begin{equation}
\mathrm{LTSim}(\mathcal{L}, \hat{\mathcal{L}})
\;=\;
\exp\,\Bigl(-\,D(\mathcal{L}, \hat{\mathcal{L}})\Bigr)
\end{equation}
A lower cost \(D(\mathcal{L}, \hat{\mathcal{L}})\) implies higher layout similarity; we set the scale parameter to \(1.0\) for simplicity \cite{21}.

To retrieve the top-\(K\) examples for a given layout \(\mathcal{L}\), we consider a database \(\mathbf{DB} = \{\hat{\mathcal{L}}_1, \hat{\mathcal{L}}_2, \dots, \hat{\mathcal{L}}_{N}\}\). We compute
\(
\mathrm{LTSim}\bigl(\mathcal{L}, \hat{\mathcal{L}}_i\bigr)
\)
for each candidate \(\hat{\mathcal{L}}_i\in\mathbf{DB}\), and select the top-\(K\) layouts that yield the highest scores:
\begin{equation}
\label{eq:topk}
\mathcal{R}_{K}(\mathcal{L})
\;=\;
\mathrm{TopK}_{\,\hat{\mathcal{L}}\in \mathbf{DB}}\,
\mathrm{LTSim}\bigl(\mathcal{L},\,\hat{\mathcal{L}}\bigr)
\end{equation}
We then collect these \(\hat{\mathcal{L}}_k\in \mathcal{R}_{K}(\mathcal{L})\) as prompting examples in subsequent generation stages. By leveraging the soft matching mechanism in \(\mathrm{LTSim}\), the Layout-aware RAG remains robust even when the input and candidate layouts differ in the number or types of elements.

\begin{table*}[t!]
  \centering
  \label{Table.2}
  \caption{\textbf{Quantitative comparison on Content-aware Layout Generation.} We report results on both PKU (top) and CGL (bottom) datasets. \textbf{LayoutCoT} (Ours) consistently outperforms baselines across most metrics. Note that $^*$ denotes variants using Gemini-2.5-pro, and $^{\dagger}$ denotes using DeepSeek-R1 without CoT.}
  \label{tab:content_aware_main}
  \resizebox{0.9\textwidth}{!}{
  \begin{tabular}{l c ccc ccccccc}
    \toprule
    \multirow{2}{*}{\textbf{Method}} & \multirow{2}{*}{Training-free}
      & \multicolumn{3}{c}{\textbf{Content}}
      & \multicolumn{6}{c}{\textbf{Graphic}} \\
    \cmidrule(lr){3-5}\cmidrule(l){6-11}
      & & Occ$\downarrow$ & Rea$\downarrow$ & Uti$\uparrow$
        & Align$\downarrow$ & Und$_l \uparrow$ & Und$_s \uparrow$
        & Ove$\downarrow$ & Val$\uparrow$ & $R_e\uparrow$ \\
    \midrule
    \multicolumn{11}{c}{\textit{\textbf{Dataset: PKU}}} \\
    \midrule
    CGL-GAN \cite{47}            & \xmark      & 0.219 & 0.175 & 0.226 & 0.0062 & 0.575 & 0.259 & 0.0605 & 0.707 & --   \\
    DS-GAN  \cite{11}            & \xmark      & 0.209 & 0.217 & 0.229 & 0.0046 & 0.585 & 0.424 & 0.0260 & 0.903 & --   \\
    RALF \cite{43}               & \xmark      & 0.171 & 0.150 & 0.246 & 0.0007 & 0.893 & 0.840 & 0.0180 & 0.998 & 0.981 \\
    LayoutPrompter \cite{12}     & \checkmark  & 0.251 & 0.171 & 0.237 & 0.0021 & 0.824 & 0.809 & 0.0005 & 0.997 & 0.865 \\
    LayoutPrompter$^*$ \cite{12} & \checkmark & 0.184 & 0.179 & 0.215 & 0.0025 & 0.799 & 0.751 & 0.0011 & 0.995 & 0.869 \\ 
    LayoutCoT$^{\dagger}$        & \checkmark  & 0.214 & 0.180 & 0.264 & 0.0016 & 0.925 & 0.849 & \textbf{0.0004} & \textbf{1.000} & 0.806 \\
    \textbf{LayoutCoT (ours)}    & \checkmark  & 0.207 & \textbf{0.074} & \textbf{0.289} & \textbf{0.0002} & \textbf{0.980} & \textbf{0.946} & 0.0013 & \textbf{1.000} & \textbf{1.000} \\
    \textbf{LayoutCoT$^*$ (ours)} & \checkmark & \textbf{0.143} & 0.139 & 0.218 & 0.0011 & 0.821 & 0.770 & 0.0024 & \textbf{1.000} & \textbf{1.000}\\
    
    \midrule
    \multicolumn{11}{c}{\textit{\textbf{Dataset: CGL}}} \\
    \midrule
    CGL-GAN \cite{47}            & \xmark      & 0.489 & 0.268 & 0.147 & 0.0420 & 0.275 & 0.244 & 0.269 & 0.876 & --   \\
    DS-GAN  \cite{11}            & \xmark      & 0.451 & 0.224 & 0.193 & 0.0485 & 0.370 & 0.301 & 0.075 & 0.893 & --   \\
    RALF \cite{43}               & \xmark      & 0.336 & 0.197 & 0.247 & 0.0023 & 0.943 & 0.884 & 0.027 & 0.995 & 0.874 \\
    LayoutPrompter \cite{12}     & \checkmark  & 0.415 & 0.184 & 0.227 & 0.0091 & 0.338 & 0.317 & 0.0049 & 0.996 & 0.812 \\
    LayoutCoT$^{\dagger}$        & \checkmark  & 0.267 & 0.182 & 0.225 & 0.0018 & 0.515 & 0.501 & 0.0023 & 0.994 & 0.803 \\
    \textbf{LayoutCoT (ours)}    & \checkmark  & \textbf{0.170} & \textbf{0.101} & \textbf{0.260} & \textbf{0.0005} & \textbf{0.958} & \textbf{0.958} & \textbf{0.0018} & \textbf{0.998} & \textbf{0.939} \\
    \bottomrule
  \end{tabular}}
\end{table*}

\subsection{Coarse Layout Generation}
\label{sec3.2}

After retrieving the top-$K$ examples via RAG (Section~\ref{sec3.1}), we convert each layout
$\hat{\mathcal{L}}_k\!\in\!\mathcal{R}_{K}(\mathcal{L})$ into a pre‑defined HTML snippet, the format definition of the HTML snippets is consistent with that of LayoutPrompter.
Let the unified prompt set be
\begin{equation}
     \mathcal{S}\;=\;
  \bigl\{
    \mathrm{HTML}(\hat{\mathcal{L}}_{1}),\,\dots,\,
    \mathrm{HTML}(\hat{\mathcal{L}}_{K}),\;
    c
  \bigr\} 
\end{equation}
where $c$ contains user‑defined constraints.
Feeding $\mathcal{S}$ to the LLM yields $n$ candidate layouts in parallel:
\begin{equation}
  \bigl\{
    \mathcal{L}^{j}_{\mathrm{HTML}}
  \bigr\}_{j=1}^{n}
  = \text{LLM}(\mathcal{S})
\end{equation}
We score each candidate with a ranker $\tau(\cdot)$ \cite{12} and
select the highest‑scoring one as the coarse result:
\begin{equation}
  \mathcal{L}^{\mathrm{coarse}}_{\mathrm{HTML}}
  \;=\;
  \arg\max_{\,1\le j\le n}\;
  \tau\bigl(\mathcal{L}^{j}_{\mathrm{HTML}}\bigr)
\end{equation}
This coarse layout preserves the user’s key structural constraints and
serves as the starting point for refinement.

\subsection{Chain of Thought for Layout Refinement}
\label{sec3.3}

Given $\mathcal{L}^{\mathrm{coarse}}_{\mathrm{HTML}}$ and the same constraint $c$,
we apply a multi‑stage \textbf{Chain‑of‑Thought} strategy
to iteratively improve the design.
Let the task be decomposed into $i$ refinement sub‑problems
$\{Q_L^{1},\dots,Q_L^{i}\}$.
Denote $\mathcal{L}^{\mathrm{refine}_{0}}_{\mathrm{HTML}}
      =\mathcal{L}^{\mathrm{coarse}}_{\mathrm{HTML}}$.
At stage $t$ ($1\!\le\!t\!\le\!3$) we query the LLM with the
augmented prompt
$\mathcal{S}\cup\{\mathcal{L}^{\mathrm{refine}_{t-1}}_{\mathrm{HTML}},\,Q_L^{t}\}$:
\begin{equation}
\resizebox{0.9\linewidth}{!}{%
$
  \mathcal{L}^{\mathrm{refine}_{t}}_{\mathrm{HTML}}
  =
  \text{LLM}\Bigl(
    \mathcal{S}\cup
    \bigl\{\mathcal{L}^{\mathrm{refine}_{t-1}}_{\mathrm{HTML}},\,
           Q_L^{t}\bigr\}
  \Bigr),
  \, t=1,2,3
$
}
\end{equation}
The final refined layout
$\mathcal{L}^{\mathrm{refine}_{3}}_{\mathrm{HTML}}$
is thus obtained after $3$ sequential reasoning steps,
yielding a design that is both semantically coherent
and visually pleasing.

\section{Experiments}
\subsection{Experimental Setup}
We conduct detailed experiments on three different types of tasks: content-aware, constraint-explicit, and text-to-layout, using five distinct datasets: PKU \cite{11}, CGL \cite{23}, RICO \cite{25}, Publaynet \cite{26}, and WebUI \cite{24}. Since no official test set splits are provided for the PKU and CGL datasets, we used 9,974 labeled images in the PKU dataset as the training set and the database for the Layout-aware RAG module, with 905 unlabeled images serving as the test set. In the CGL dataset, we took 38,510 images as the training set and the database for the Layout-aware RAG module, and selected an additional 1,647 images (with no overlap with the training set) as the test set. For other datasets, we adopt the official data splits, as detailed in Table .\ref{Table.1}

\subsection{Implement Details}
In our experiments, we provide details on the choice of the 
$k$-value for the top-$k$ retrieval in the Layout-aware RAG. When generating the Coarse Layout, we follow the approach of LayoutPrompter \cite{12} and set $k$=10. For the CoT stage, we set $k$=4. We also adopt LayoutPrompter’s setting of $n$=10 for the number of generation times by LLMs when selecting the Coarse Layout. Since the 'Text-Davinci-003' model used by LayoutPrompter is no longer available, we replaced the LLM with 'gpt-4' in our comparative experiments. Additionally, on the PKU dataset, we conducted extra experiments using Gemini-2.5-pro, denoted as \textbf{LayoutPrompter$^*$}. In the LayoutCoT experiments, we used 'gpt-4' as the LLM. To further validate the effectiveness of our CoT module, we provided \textbf{LayoutCoT$^\dag$} results for each experiment; \textbf{LayoutCoT$^\dag$} refers to the variant where the LLM is replaced by the reasoning model 'DeepSeek-R1' and the CoT module is removed (only LayoutAware RAG and Coarse Layout Generation are used). For the PKU dataset, we also conducted experiments using Gemini-2.5-pro, denoted as \textbf{LayoutCoT$^*$}.

\begin{figure}[!htbp]
    \centering
    \includegraphics[width=1.0\linewidth]{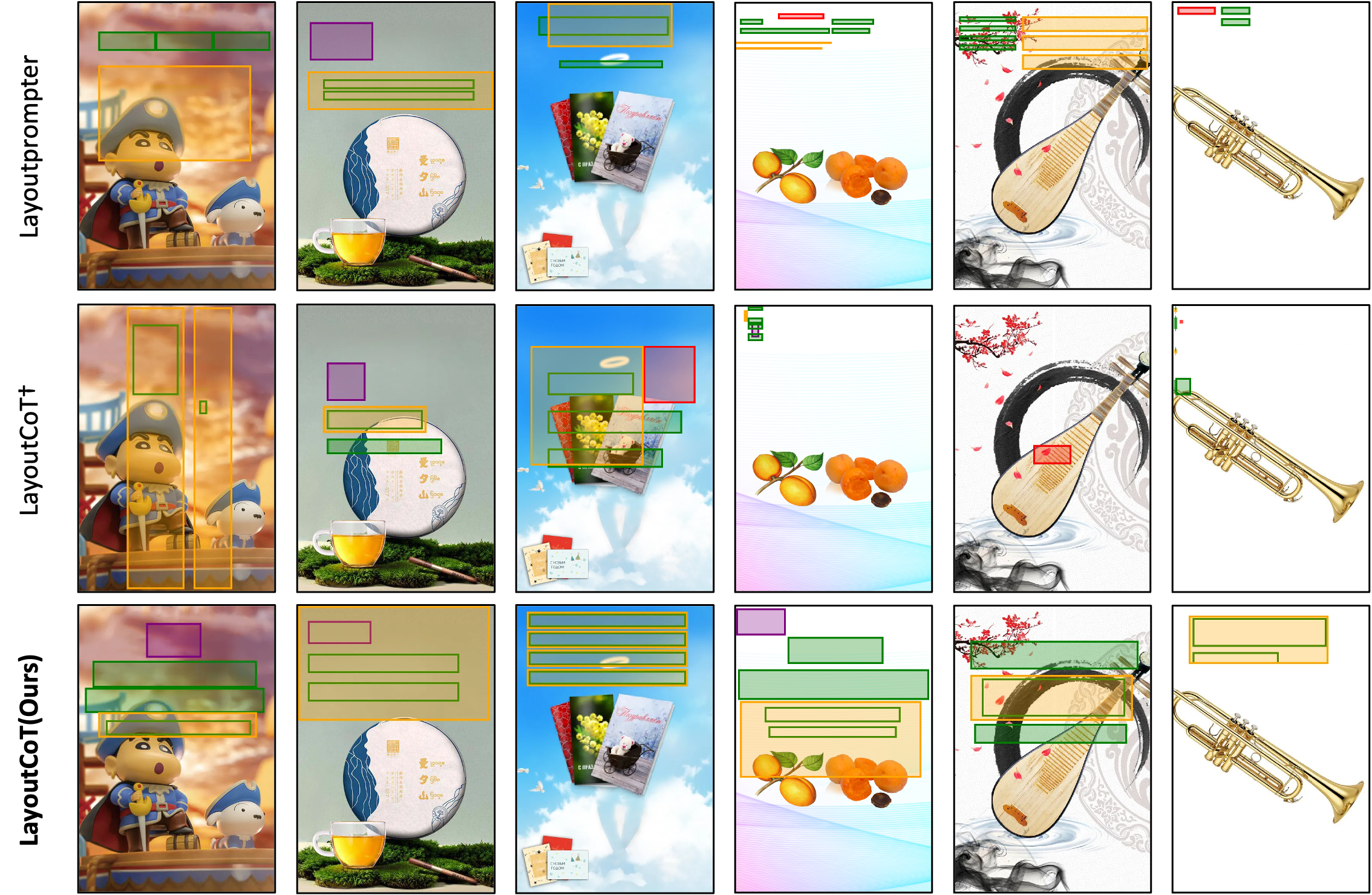}
    \caption{Qualitative Results for the Content-aware Layout Generation Task. LayoutPrompter and LayoutCoT$^\dag$ tend to generate dense, small boxes in the upper-left corner, which is unreasonable. In contrast, LayoutCoT effectively corrects this error and achieves satisfactory results.}
    \label{Fig.4}
\end{figure}

\subsection{Evaluation Metrics}
\label{Sec4.3}
For the constraint-explicit layout generation and text-to-layout tasks, we utilize four standard metrics for evaluation: \textbf{Alignment} measures how well the elements within the layout align with each other. \textbf{Overlap} calculates the area of overlap between any two elements in the layout. \textbf{Maximum IoU, mIoU} assesses the maximum intersection over union between the generated layout and the actual layout, providing a measure of layout accuracy. \textbf{Frechet Inception Distance (FID)} quantifies the similarity between the distribution of generated layouts and the distribution of real layouts, using features extracted from a deep neural network trained on images. For the content-aware layout generation task, we reference the eight metrics used in \cite{11} along with a newly designed metric $R_e$ that we introduce. The metric $R_e$ is used to assess the reasonableness of each element's size within the layout. Let 
\[
  r_i = \frac{S^{i}}{S^{i}_{\text{train}}},\qquad
  d_i = \bigl|\ln r_i\bigr|,\qquad
  \tau = \ln(1.1)
\]
where \(S^{i}\) is the predicted mean area of label~\(i\) and
\(S^{i}_{\text{train}}\) is its training-set reference.
With a \(\pm10\%\) tolerance band \(\tau\), we define
\begin{equation}
  \operatorname{score}_i
  = \exp\,\!\bigl(-\max(0,\, d_i - \tau)\bigr)
  \label{eq:score}
\end{equation}
Thus \(\operatorname{score}_i = 1\) when \(0.9 \le r_i \le 1.1\) and decays exponentially once the deviation exceeds \(10\%\).

Aggregating over all \(N\) labels yields
\begin{equation}
  R_e
  = \exp \Bigl(
      -\sqrt{1/N \sum\nolimits_{i=1}^{N}
        \bigl(\max(0,\, d_i - \tau)\bigr)^{2}}
    \Bigr)
  \label{eq:Re_rms}
\end{equation}
where \(R_e \in (0,1]\); higher values indicate more reasonable element sizes.

\subsection{Results on Content-Aware Layout Generation}
\label{Sec4.4}
The quantitative experimental results for both PKU and CGL datasets are summarized in Table \ref{tab:content_aware_main}.. As seen from the table, our method surpasses all the evaluation metrics used in content-aware layout generation tasks compared to both the training-based baseline methods and another training-free method, LayoutPrompter \cite{12}. Notably, our proposed CoT module enables "GPT-4" to even outperform the deep reasoning large language model "DeepSeek-R1" in this layout generation task.
Moreover, our performance on the Overlap metric in the PKU dataset was suboptimal, prompting further visual analysis. As illustrated in Figure. \ref{Fig.4}, the very low scores on the Overlap metric for both LayoutPrompter and LayoutCoT$^{\dagger}$ result from numerous bad cases. In these cases, layout bounding boxes are extremely small and densely clustered in the upper-left corner, leading to low Overlap values but highly unaesthetic and impractical layout outcomes. Hence, we introduce the $R_e$ metric in Sec. \ref{Sec4.3} to measure the reasonableness of layout sizes, on which we have performed exceptionally well. Qualitative comparative cases are further provided in Figure. \ref{Fig.4}.

\begin{table}[t!]
    \centering
    \caption{Quantitative comparison with other methods on constraint-explicit layout generation tasks.}
    \label{Table.3}
    \setlength{\tabcolsep}{1.2pt} 
    \renewcommand{\arraystretch}{1.1}
    \resizebox{\linewidth}{!}{
        \begin{tabular}{llcccccccc}
            \toprule
            &  & \multicolumn{4}{c}{RICO}  & \multicolumn{4}{c}{PubLayNet} \\ 
            \cmidrule(lr){3-6}  \cmidrule(lr){7-10}
            Tasks  & Methods & mIoU$\uparrow$ & FID$\downarrow$ & Align.$\downarrow$ & Over.$\downarrow$ & mIoU$\uparrow$ 
            & FID$\downarrow$ & Align.$\downarrow$ & Over.$\downarrow$ \\ \hline
            \multirow{5}{*}{Gen-T} & BLT \cite{4} & 0.216 & 25.63 & 0.150 &  0.983 & 0.140 & 38.68 & 0.036 & 0.196 \\ 
            & LayoutFormer++ \cite{2} & 0.432 & \textbf{1.096} & 0.230 & 0.530  & 0.348 & 8.411 & 0.020 & 0.008  \\ 
            & LayoutPrompter \cite{12} & 0.604 & 2.134 & 0.002 & 0.029  & 0.619 & 1.648 & 0.001 & 0.004  \\ 
            & LayoutCoT$^{\dagger}$ & 0.718 & 4.970 & 0.002 & \textbf{0.021} & \textbf{0.795} & 2.608 & \textbf{0.000} & 0.003\\
            & \textbf{LayoutCoT (Ours)} & \textbf{0.719} & 2.189 & \textbf{0.001} & 0.022  & 0.772 & \textbf{1.227} & \textbf{0.000} & \textbf{0.002}  \\
            \hline
            \multirow{5}{*}{Gen-TS} & BLT \cite{4} & 0.604 & 0.951 & 0.181 & 0.660  & 0.428 & 7.914 & 0.021 & 0.419  \\
            & LayoutFormer++ \cite{2} & 0.620 & \textbf{0.757} & 0.202 & 0.542  & 0.471 & 0.720 & 0.024 & 0.037  \\
            & LayoutPrompter \cite{12} & 0.707 & 3.072 & 0.002 & 0.051  & 0.727 & 1.863 & 0.001 & 0.076\\ 
            & LayoutCoT$^{\dagger}$ & 0.782 & 1.252 & 0.001 & 0.033 & 0.844 & 1.701 & \textbf{0.000} & 0.034 \\
            & \textbf{LayoutCoT (Ours)} & \textbf{0.795} & 3.506 & \textbf{0.000} & \textbf{0.031} & \textbf{0.847} & \textbf{1.430} & \textbf{0.000} & \textbf{0.004} \\
            \hline
            \multirow{5}{*}{Gen-R} & CLG-LO \cite{3} & 0.286 & 8.898 & 0.311 & 0.615  & 0.277 & 19.73 & 0.123 & 0.200  \\
            & LayoutFormer++ \cite{2} & 0.424 & 5.972 & 0.332 & 0.537 & 0.353 & 4.954 & 0.025 & 0.076 \\
            & LayoutPrompter \cite{12} & 0.583 & \textbf{2.249} & \textbf{0.001} & 0.044 & 0.627 & 2.224 & 0.001 & 0.011 \\ 
            & LayoutCoT$^{\dagger}$ & 0.724 & 2.399 & \textbf{0.001} & 0.030 & \textbf{0.779} & 1.853 & \textbf{0.000} & 0.004 \\
            & \textbf{LayoutCoT (Ours)} & \textbf{0.725} & 3.569 & \textbf{0.001} & \textbf{0.015} & 0.769 & \textbf{1.408} & 0.001 & \textbf{0.002} \\
            \hline
            \multirow{5}{*}{Comp.} & LayoutTrans \cite{1} & 0.363 & 6.679 & 0.194 & 0.478  & 0.077 & 14.76 & 0.019 & 0.001  \\
            & LayoutFormer++ \cite{2} & \textbf{0.732} & 4.574 & 0.077 & 0.487 & 0.471 & 10.25 & 0.020 & 0.002 \\
            & LayoutPrompter \cite{12} & 0.541 & 2.656 & 0.001 & 0.041 & 0.754 & 1.282 & 0.001 & 0.001 \\ 
            & LayoutCoT$^{\dagger}$ & 0.682 & 2.236 & 0.008 & 0.031 & \textbf{0.843} & 0.840 & 0.001 & \textbf{0.000} \\
            & \textbf{LayoutCoT (Ours)} & 0.716 & \textbf{2.112} & \textbf{0.001} & \textbf{0.012} & 0.770 & \textbf{0.531} & \textbf{0.000} & 0.001 \\
            \hline
            \multirow{5}{*}{Refine} & RUITE \cite{32} & 0.811 & 0.107 & 0.133 & 0.483 & 0.781 & 0.061 & 0.029 & 0.020 \\
            & LayoutFormer++ \cite{2} & 0.816 & \textbf{0.032} & 0.123 & 0.489 & 0.785 & 0.086 & 0.024 & 0.006 \\
            & LayoutPrompter \cite{12} & 0.874 & 0.225 & 0.001 & 0.110 & 0.756 & 0.143 & 0.003 & 0.001 \\
            & LayoutCoT$^{\dagger}$ & 0.869 & 0.876 & 0.001 & 0.071 & 0.801 & 0.265 & 0.001 & \textbf{0.000} \\
            & \textbf{LayoutCoT (Ours)} & \textbf{0.890} & 0.353 & \textbf{0.000} & \textbf{0.052} & \textbf{0.839} & \textbf{0.141} & \textbf{0.000} & \textbf{0.000} \\
            \bottomrule
        \end{tabular}
    }
\end{table}

\begin{figure}[t!]
    \centering
    \includegraphics[width=0.9\linewidth]{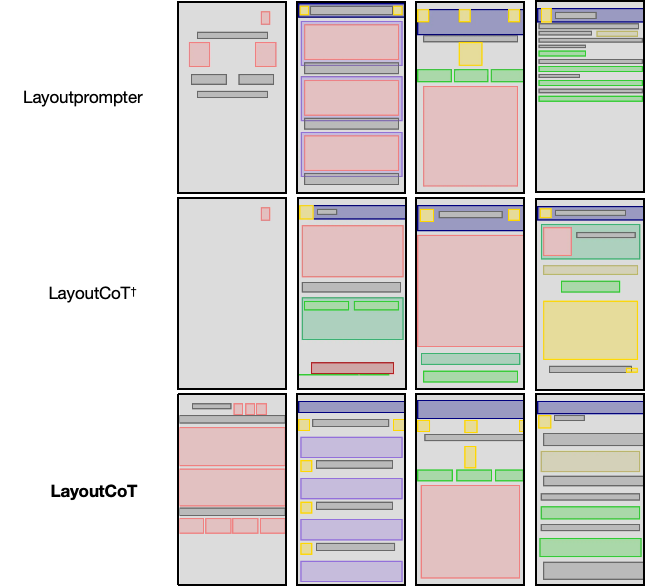}
    \caption{Qualitative Results for the Constraint-Explicit Layout Generation Task. The RICO dataset features a wide variety of label categories, making the completion task on RICO more challenging compared to other task types, we conduct visualizations based on the completion task of the RICO dataset. It is evident that LayoutCoT designs more rational layouts, with better element overlap and alignment compared to other methods.}
    \label{Fig.5}
\end{figure}

\subsection{Results on Constraint-Explicit Layout Generation}
Table \ref{Table.3} presents the quantitative evaluation comparing LayoutCoT with several training-based baselines and the training-free method LayoutPrompter. From the metrics, it is evident that, even without any training, LayoutCoT achieves optimal performance on most indicators. The corresponding qualitative results in Figure \ref{Fig.5} demonstrate that LayoutCoT exhibits outstanding layout control capabilities and generation quality. LayoutCoT not only effectively meets various input constraints--including element type, size, and relationship constraints--but also produces visually pleasing layouts by avoiding overlap between elements and enhancing overall alignment. These findings validate the effectiveness of LayoutCoT.

\subsection{Results on Text-to-Layout}
The quantitative metrics are presented in Table \ref{Table.4}. Because text-to-layout is one of the most challenging tasks in layout generation, LayoutCoT falls short in the mIoU metric compared to the parse-then-place \cite{29} method. However, it performs exceptionally well on the remaining three metrics. In many scenarios, LayoutCoT more effectively fulfills textual descriptions and excels in aesthetics, avoiding element overlap and maintaining element alignment.

\begin{table}[!htbp]
    \centering
    \caption{Quantitative comparison with baselines on text-to-layout. We compare our method with the baselines Mockup \cite{28}, parse-then-place \cite{29}, and LayoutPrompter.}
    \renewcommand{\arraystretch}{1.2}
\resizebox{0.82\linewidth}{!}{
    \begin{tabular}{lcccc}
        \toprule
        Methods &  mIoU  $\uparrow$  & FID  $\downarrow$  & Align. $\downarrow$  & Overlap $\downarrow$  \\ \midrule
        Mockup & 0.193 & 37.012 & 0.0059 & 0.4348 \\
        parse-then-place & \textbf{0.684} & 2.959 & 0.0008 & 0.1380\\
        LayoutPrompter & 0.174 & 4.773 & 0.0009 & 0.0107 \\
        \midrule
        LayoutCoT$^{\dagger}$ & 0.164 & \textbf{1.018} & 0.0009 & 0.0068 \\
        \textbf{LayoutCoT (Ours)} & 0.199 & 1.732 & \textbf{0.00006} & \textbf{0.0061} \\
        \bottomrule
    \end{tabular}
}
    \label{Table.4}
\end{table}

\subsection{Ablation Studies}
We validate the effectiveness of each component on the PKU dataset, as summarized in Table~\ref{tab:all_ablations}.

\noindent \textbf{Impact of Modules.} 
Table~\ref{tab:all_ablations} (a) demonstrates that both the Layout-aware RAG and CoT modules individually improve performance over the baseline. Specifically, RAG enhances structural alignment (Align $0.0021\!\rightarrow\!0.0005$), while CoT effectively reduces occlusion. The full LayoutCoT framework achieves the best balance across all metrics, confirming the complementarity of retrieval context and reasoning capabilities. The slight rise in Overlap is a necessary trade-off to correct the baseline's tendency to generate unrealistically small, crowded boxes.

\noindent \textbf{Retrieval Strategy Analysis.} 
Regarding the RAG module, Table~\ref{tab:all_ablations} (b) shows that our LTSim-based retrieval significantly outperforms Random, mIoU, and DreamSim strategies by effectively capturing holistic layout similarity. Furthermore, we investigate the number of retrieved examples $k$ in Table~\ref{tab:all_ablations} (c), finding that $k=4$ offers the optimal trade-off between providing sufficient context and minimizing noise.

\noindent \textbf{CoT Design Validation.} 
Finally, we analyze the reasoning mechanism. Table~\ref{tab:all_ablations} (d) reveals that our multi-stage strategy markedly surpasses the single-stage approach, validating the necessity of decomposing complex layout tasks into logical steps. Moreover, Table~\ref{tab:all_ablations} (e) confirms that our domain-specific manually designed CoT outperforms the generic Self-Refine mechanism in 7 out of 8 metrics, ensuring superior layout rationality.

\begin{table}[t!]
  \centering
  \caption{\textbf{Comprehensive Ablation Studies on the PKU dataset.} We systematically analyze the impact of (a) overall modules, (b) layout-aware retrieval strategies, (c) the number of retrieved examples ($k$), (d) reasoning stages, and (e) refinement methodologies. The best results are highlighted in \textbf{bold}.}
  \label{tab:all_ablations}
  \resizebox{\linewidth}{!}{
  \begin{tabular}{l cccccccc}
    \toprule
    \textbf{Settings} & \textbf{Occ}$\downarrow$ & \textbf{Rea}$\downarrow$ & \textbf{Uti}$\uparrow$ & \textbf{Align}$\downarrow$ & \textbf{Und$_l$}$\uparrow$ & \textbf{Und$_s$}$\uparrow$ & \textbf{Ove}$\downarrow$ & \textbf{Val}$\uparrow$ \\
    \midrule
    
    \multicolumn{9}{l}{\textit{\textbf{(a) Overall Module Ablation}}} \\
    Baseline (No RAG/CoT) & 0.251 & 0.171 & 0.237 & 0.0021 & 0.824 & 0.809 & \textbf{0.0005} & 0.997 \\
    w/ RAG only           & 0.219 & 0.167 & 0.241 & 0.0005 & 0.933 & 0.851 & 0.0017 & 0.999 \\
    w/ CoT only           & 0.209 & 0.161 & 0.231 & 0.0004 & 0.931 & 0.844 & 0.0014 & 0.999 \\
    \textbf{LayoutCoT (Full)}      & \textbf{0.207} & \textbf{0.074} & \textbf{0.289} & \textbf{0.0002} & \textbf{0.980} & \textbf{0.946} & 0.0013 & \textbf{1.000} \\
    \midrule

    \multicolumn{9}{l}{\textit{\textbf{(b) Layout-aware RAG Strategy}}} \\
    No RAG                & 0.209 & 0.161 & 0.231 & 0.0004 & 0.931 & 0.844 & 0.0014 & 0.999 \\
    Random                & 0.299 & 0.184 & 0.241 & 0.0009 & 0.808 & 0.783 & 0.0023 & 0.997 \\
    mIoU                  & 0.230 & 0.163 & 0.265 & 0.0003 & 0.951 & 0.894 & 0.0010 & \textbf{1.000} \\
    DreamSim              & 0.252 & 0.086 & 0.273 & 0.0004 & 0.972 & 0.932 & \textbf{0.0008} & 0.999 \\
    \textbf{LTSim (Ours)} & \textbf{0.207} & \textbf{0.074} & \textbf{0.289} & \textbf{0.0002} & \textbf{0.980} & \textbf{0.946} & 0.0013 & \textbf{1.000} \\
    \midrule

    \multicolumn{9}{l}{\textit{\textbf{(c) Number of Retrieved Examples ($k$)}}} \\
    $k=2$                 & 0.232 & 0.119 & 0.282 & 0.0009 & 0.835 & 0.797 & 0.0019 & 0.999 \\
    \textbf{$k=4$ (Ours)} & \textbf{0.207} & \textbf{0.074} & 0.289 & \textbf{0.0002} & \textbf{0.980} & \textbf{0.946} & 0.0013 & \textbf{1.000} \\
    $k=6$                 & 0.216 & 0.075 & \textbf{0.301} & 0.0004 & 0.968 & 0.939 & \textbf{0.0010} & \textbf{1.000} \\
    \midrule

    \multicolumn{9}{l}{\textit{\textbf{(d) CoT Reasoning Stages}}} \\
    Single-Stage (1)      & 0.219 & 0.167 & 0.241 & 0.0005 & 0.933 & 0.851 & 0.0017 & 0.999 \\
    \textbf{Multi-Stage (3)}       & \textbf{0.207} & \textbf{0.074} & \textbf{0.289} & \textbf{0.0002} & \textbf{0.980} & \textbf{0.946} & \textbf{0.0013} & \textbf{1.000} \\
    \midrule

    \multicolumn{9}{l}{\textit{\textbf{(e) Refinement Method}}} \\
    Self-Refine           & \textbf{0.153} & 0.149 & 0.213 & 0.0012 & 0.868 & 0.806 & 0.0029 & \textbf{1.000} \\
    \textbf{Manual CoT (Ours)}     & 0.207 & \textbf{0.074} & \textbf{0.289} & \textbf{0.0002} & \textbf{0.980} & \textbf{0.946} & \textbf{0.0013} & \textbf{1.000} \\
    
    \bottomrule
  \end{tabular}
  }
  \label{Table.5}
\end{table}

\section{Conclusion}
This paper focuses on unleashing the deep reasoning abilities of LLMs in layout generation tasks without requiring any training. To address the limitations of existing methods and enhance models’ performance in layout generation, we propose LayoutCoT, which comprises three key components: a Layout-aware RAG, a Coarse Layout Generator, and a Chain of Thought for Layout Refinement. Specifically, we select prompting examples through a carefully designed RAG approach to generate a coarse layout, then further decompose the layout task into multiple simpler subproblems. By integrating the RAG results, we refine the coarse layout. Our method activates the deep reasoning capabilities of LLMs in layout generation, achieving state-of-the-art results on five public datasets spanning three different types of layout tasks. These findings demonstrate that LayoutCoT is a highly versatile, data-efficient, and training-free approach that produces high-quality layouts while satisfying given constraints.

\section{Limitations}
Our work focuses on unlocking the reasoning potential of LLMs, but this comes with trade-offs. The iterative nature of our CoT mechanism, while boosting performance, incurs higher computational overhead than non-autoregressive models. Additionally, since our method leverages in-context learning, its upper bound is influenced by the retrieval database; expanding the database with more diverse layouts would be necessary to handle extremely rare or complex constraints effectively.

\bibliography{anthology}

@inproceedings{1,
  title={LayoutTransformer: Layout generation and completion with self-attention},
  author={Gupta, Kamal and Lazarow, Justin and Achille, Alessandro and Davis, Larry S and Mahadevan, Vijay and Shrivastava, Abhinav},
  booktitle={Proceedings of the IEEE/CVF International Conference on Computer Vision (ICCV)},
  pages={1004--1014},
  year={2021}
}

@article{2,
  title={UniLayout: Taming unified sequence-to-sequence transformers for graphic layout generation},
  author={Jiang, Zhaoyun and Deng, Huayu and Wu, Zhongkai and Guo, Jiaqi and Sun, Shizhao and Mijovic, Vuksan and Yang, Zijiang and Lou, Jian-Guang and Zhang, Dongmei},
  journal={arXiv preprint arXiv:2208.08037},
  year={2022}
}

@inproceedings{3,
  title={Constrained graphic layout generation via latent optimization},
  author={Kikuchi, Kotaro and Simo-Serra, Edgar and Otani, Mayu and Yamaguchi, Kota},
  booktitle={Proceedings of the 29th ACM International Conference on Multimedia},
  pages={88--96},
  year={2021}
}

@inproceedings{4,
  title={{BLT}: Bidirectional layout transformer for controllable layout generation},
  author={Kong, Xiang and Jiang, Lu and Chang, Huiwen and Zhang, Han and Hao, Yuan and Gong, Haifeng and Essa, Irfan},
  booktitle={European Conference on Computer Vision (ECCV)},
  pages={474--490},
  year={2022},
  organization={Springer}
}

@inproceedings{5,
  title={Neural design network: Graphic layout generation with constraints},
  author={Lee, Hsin-Ying and Jiang, Lu and Essa, Irfan and Le, Phuong B and Gong, Haifeng and Yang, Ming-Hsuan and Yang, Weilong},
  booktitle={European Conference on Computer Vision (ECCV)},
  pages={491--506},
  year={2020},
  organization={Springer}
}

@article{6,
  title={Attribute-conditioned layout {GAN} for automatic graphic design},
  author={Li, Jianan and Yang, Jimei and Zhang, Jianming and Liu, Chang and Wang, Christina and Xu, Tingfa},
  journal={IEEE Transactions on Visualization and Computer Graphics},
  volume={27},
  number={10},
  pages={4039--4048},
  year={2020},
  publisher={IEEE}
}

@article{7,
  title={LayoutGAN: Generating graphic layouts with wireframe discriminators},
  author={Li, Jianan and Yang, Jimei and Hertzmann, Aaron and Zhang, Jianming and Xu, Tingfa},
  journal={arXiv preprint arXiv:1901.06767},
  year={2019}
}

@inproceedings{8,
  title={LayoutDiffusion: Controllable diffusion model for layout-to-image generation},
  author={Zheng, Guangcong and Zhou, Xianpan and Li, Xuewei and Qi, Zhongang and Shan, Ying and Li, Xi},
  booktitle={Proceedings of the IEEE/CVF Conference on Computer Vision and Pattern Recognition (CVPR)},
  pages={22490--22499},
  year={2023}
}

@inproceedings{9,
  title={Unifying layout generation with a decoupled diffusion model},
  author={Hui, Mude and Zhang, Zhizheng and Zhang, Xiaoyi and Xie, Wenxuan and Wang, Yuwang and Lu, Yan},
  booktitle={Proceedings of the IEEE/CVF Conference on Computer Vision and Pattern Recognition (CVPR)},
  pages={1942--1951},
  year={2023}
}

@inproceedings{10,
  title={LayoutDM: Discrete diffusion model for controllable layout generation},
  author={Inoue, Naoto and Kikuchi, Kotaro and Simo-Serra, Edgar and Otani, Mayu and Yamaguchi, Kota},
  booktitle={Proceedings of the IEEE/CVF Conference on Computer Vision and Pattern Recognition (CVPR)},
  pages={10167--10176},
  year={2023}
}

@inproceedings{11,
  title={PosterLayout: A new benchmark and approach for content-aware visual-textual presentation layout},
  author={Hsu, Hsiao Yuan and He, Xiangteng and Peng, Yuxin and Kong, Hao and Zhang, Qing},
  booktitle={Proceedings of the IEEE/CVF Conference on Computer Vision and Pattern Recognition (CVPR)},
  pages={6018--6026},
  year={2023}
}

@article{12,
  title={LayoutPrompter: Awaken the design ability of large language models},
  author={Lin, Jiawei and Guo, Jiaqi and Sun, Shizhao and Yang, Zijiang and Lou, Jian-Guang and Zhang, Dongmei},
  journal={Advances in Neural Information Processing Systems (NeurIPS)},
  volume={36},
  pages={43852--43879},
  year={2023}
}

@article{13,
  title={Generative adversarial nets},
  author={Goodfellow, Ian J and Pouget-Abadie, Jean and Mirza, Mehdi and Xu, Bing and Warde-Farley, David and Ozair, Sherjil and Courville, Aaron and Bengio, Yoshua},
  journal={Advances in Neural Information Processing Systems (NeurIPS)},
  volume={27},
  year={2014}
}

@article{14,
  title={Denoising diffusion probabilistic models},
  author={Ho, Jonathan and Jain, Ajay and Abbeel, Pieter},
  journal={Advances in Neural Information Processing Systems (NeurIPS)},
  volume={33},
  pages={6840--6851},
  year={2020}
}

@article{15,
  title={Language models are few-shot learners},
  author={Brown, Tom and Mann, Benjamin and Ryder, Nick and Subbiah, Melanie and Kaplan, Jared D and Dhariwal, Prafulla and Neelakantan, Arvind and Shyam, Pranav and Sastry, Girish and Askell, Amanda and others},
  journal={Advances in Neural Information Processing Systems (NeurIPS)},
  volume={33},
  pages={1877--1901},
  year={2020}
}

@misc{16,
  title={Mathematical reasoning using large language models},
  author={Imani, Shima and Shrivastava, Harsh and Du, Liang},
  year={2024},
  month={September},
  note={US Patent App. 18/144,802}
}

@article{17,
  title={{GPT-4} technical report},
  author={Achiam, Josh and Adler, Steven and Agarwal, Sandhini and Ahmad, Lama and Akkaya, Ilge and Aleman, Florencia Leoni and Almeida, Diogo and Altenschmidt, Janko and Altman, Sam and Anadkat, Shyamal and others},
  journal={arXiv preprint arXiv:2303.08774},
  year={2023}
}

@article{18,
  title={{DOC}: Improving long story coherence with detailed outline control},
  author={Yang, Kevin and Klein, Dan and Peng, Nanyun and Tian, Yuandong},
  journal={arXiv preprint arXiv:2212.10077},
  year={2022}
}

@article{19,
  title={{Re3}: Generating longer stories with recursive reprompting and revision},
  author={Yang, Kevin and Tian, Yuandong and Peng, Nanyun and Klein, Dan},
  journal={arXiv preprint arXiv:2210.06774},
  year={2022}
}

@article{20,
  title={Evaluating large language models trained on code},
  author={Chen, Mark and Tworek, Jerry and Jun, Heewoo and Yuan, Qiming and Pinto, Henrique Ponde De Oliveira and Kaplan, Jared and Edwards, Harri and Burda, Yuri and Joseph, Nicholas and Brockman, Greg and others},
  journal={arXiv preprint arXiv:2107.03374},
  year={2021}
}

@article{21,
  title={{LTSim}: Layout Transportation-based Similarity Measure for Evaluating Layout Generation},
  author={Otani, Mayu and Inoue, Naoto and Kikuchi, Kotaro and Togashi, Riku},
  journal={arXiv preprint arXiv:2407.12356},
  year={2024}
}

@article{22,
  title={{DeepSeek-R1}: Incentivizing reasoning capability in {LLMs} via reinforcement learning},
  author={Guo, Daya and Yang, Dejian and Zhang, Haowei and Song, Junxiao and Zhang, Ruoyu and Xu, Runxin and Zhu, Qihao and Ma, Shirong and Wang, Peiyi and Bi, Xiao and others},
  journal={arXiv preprint arXiv:2501.12948},
  year={2025}
}

@article{23,
  title={Composition-aware graphic layout {GAN} for visual-textual presentation designs},
  author={Zhou, Min and Xu, Chenchen and Ma, Ye and Ge, Tiezheng and Jiang, Yuning and Xu, Weiwei},
  journal={arXiv preprint arXiv:2205.00303},
  year={2022}
}

@inproceedings{24,
  title={{WebUI}: A dataset for enhancing visual {UI} understanding with web semantics},
  author={Wu, Jason and Wang, Siyan and Shen, Siman and Peng, Yi-Hao and Nichols, Jeffrey and Bigham, Jeffrey P},
  booktitle={Proceedings of the 2023 CHI Conference on Human Factors in Computing Systems},
  pages={1--14},
  year={2023}
}

@inproceedings{25,
  title={Learning design semantics for mobile apps},
  author={Liu, Thomas F and Craft, Mark and Situ, Jason and Yumer, Ersin and Mech, Radomir and Kumar, Ranjitha},
  booktitle={Proceedings of the 31st Annual ACM Symposium on User Interface Software and Technology (UIST)},
  pages={569--579},
  year={2018}
}

@inproceedings{26,
  title={{PubLayNet}: Largest dataset ever for document layout analysis},
  author={Zhong, Xu and Tang, Jianbin and Yepes, Antonio Jimeno},
  booktitle={International Conference on Document Analysis and Recognition (ICDAR)},
  pages={1015--1022},
  year={2019},
  organization={IEEE}
}

@article{27,
  title={Content-aware generative modeling of graphic design layouts},
  author={Zheng, Xinru and Qiao, Xiaotian and Cao, Ying and Lau, Rynson WH},
  journal={ACM Transactions on Graphics (TOG)},
  volume={38},
  number={4},
  pages={1--15},
  year={2019},
  publisher={ACM New York, NY, USA}
}

@article{28,
  title={Creating user interface mock-ups from high-level text descriptions with deep-learning models},
  author={Huang, Forrest and Li, Gang and Zhou, Xin and Canny, John F and Li, Yang},
  journal={arXiv preprint arXiv:2110.07775},
  year={2021}
}

@inproceedings{29,
  title={A parse-then-place approach for generating graphic layouts from textual descriptions},
  author={Lin, Jiawei and Guo, Jiaqi and Sun, Shizhao and Xu, Weijiang and Liu, Ting and Lou, Jian-Guang and Zhang, Dongmei},
  booktitle={Proceedings of the IEEE/CVF International Conference on Computer Vision (ICCV)},
  pages={23622--23631},
  year={2023}
}

@inproceedings{30,
  title={{LayoutVAE}: Stochastic scene layout generation from a label set},
  author={Jyothi, Akash Abdu and Durand, Thibaut and He, Jiawei and Sigal, Leonid and Mori, Greg},
  booktitle={Proceedings of the IEEE/CVF International Conference on Computer Vision (ICCV)},
  pages={9895--9904},
  year={2019}
}

@article{31,
  title={Attention is all you need},
  author={Vaswani, Ashish and Shazeer, Noam and Parmar, Niki and Uszkoreit, Jakob and Jones, Llion and Gomez, Aidan N and Kaiser, {\L}ukasz and Polosukhin, Illia},
  journal={Advances in Neural Information Processing Systems (NeurIPS)},
  volume={30},
  year={2017}
}

@inproceedings{32,
  title={{RUITE}: Refining {UI} layout aesthetics using transformer encoder},
  author={Rahman, Soliha and Sermuga Pandian, Vinoth Pandian and Jarke, Matthias},
  booktitle={Companion Proceedings of the 26th International Conference on Intelligent User Interfaces (IUI)},
  pages={81--83},
  year={2021}
}

@article{33,
  title={{PaLM}: Scaling language modeling with pathways},
  author={Chowdhery, Aakanksha and Narang, Sharan and Devlin, Jacob and Bosma, Maarten and Mishra, Gaurav and Roberts, Adam and Barham, Paul and Chung, Hyung Won and Sutton, Charles and Gehrmann, Sebastian and others},
  journal={Journal of Machine Learning Research},
  volume={24},
  number={240},
  pages={1--113},
  year={2023}
}

@article{34,
  title={{LLaMA}: Open and efficient foundation language models},
  author={Touvron, Hugo and Lavril, Thibaut and Izacard, Gautier and Martinet, Xavier and Lachaux, Marie-Anne and Lacroix, Timoth{\'e}e and Rozi{\`e}re, Baptiste and Goyal, Naman and Hambro, Eric and Azhar, Faisal and others},
  journal={arXiv preprint arXiv:2302.13971},
  year={2023}
}

@article{35,
  title={{LayoutNUWA}: Revealing the hidden layout expertise of large language models},
  author={Tang, Zecheng and Wu, Chenfei and Li, Juntao and Duan, Nan},
  journal={arXiv preprint arXiv:2309.09506},
  year={2023}
}

@article{36,
  title={{Code Llama}: Open foundation models for code},
  author={Roziere, Baptiste and Gehring, Jonas and Gloeckle, Fabian and Sootla, Sten and Gat, Itai and Tan, Xiaoqing Ellen and Adi, Yossi and Liu, Jingyu and Sauvestre, Romain and Remez, Tal and others},
  journal={arXiv preprint arXiv:2308.12950},
  year={2023}
}

@article{37,
  title={{PosterLlama}: Bridging design ability of language model to contents-aware layout generation},
  author={Seol, Jaejung and Kim, Seojun and Yoo, Jaejun},
  journal={arXiv preprint arXiv:2404.00995},
  year={2024}
}

@article{38,
  title={{MiniGPT-4}: Enhancing vision-language understanding with advanced large language models},
  author={Zhu, Deyao and Chen, Jun and Shen, Xiaoqian and Li, Xiang and Elhoseiny, Mohamed},
  journal={arXiv preprint arXiv:2304.10592},
  year={2023}
}

@article{39,
  title={{LayoutGPT}: Compositional visual planning and generation with large language models},
  author={Feng, Weixi and Zhu, Wanrong and Fu, Tsu-jui and Jampani, Varun and Akula, Arjun and He, Xuehai and Basu, Sugato and Wang, Xin Eric and Wang, William Yang},
  journal={Advances in Neural Information Processing Systems (NeurIPS)},
  volume={36},
  pages={18225--18250},
  year={2023}
}

@article{40,
  title={Retrieval-augmented generation for knowledge-intensive {NLP} tasks},
  author={Lewis, Patrick and Perez, Ethan and Piktus, Aleksandra and Petroni, Fabio and Karpukhin, Vladimir and Goyal, Naman and K{\"u}ttler, Heinrich and Lewis, Mike and Yih, Wen-tau and Rockt{\"a}schel, Tim and others},
  journal={Advances in Neural Information Processing Systems (NeurIPS)},
  volume={33},
  pages={9459--9474},
  year={2020}
}

@article{41,
  title={{Invar-RAG}: Invariant {LLM}-aligned Retrieval for Better Generation},
  author={Liu, Ziwei and Zhang, Liang and Li, Qian and Wu, Jianghua and Zhu, Guangxu},
  journal={arXiv preprint arXiv:2411.07021},
  year={2024}
}

@article{42,
  title={A Survey on Knowledge-Oriented Retrieval-Augmented Generation},
  author={Cheng, Mingyue and Luo, Yucong and Ouyang, Jie and Liu, Qi and Liu, Huijie and Li, Li and Yu, Shuo and Zhang, Bohou and Cao, Jiawei and Ma, Jie and others},
  journal={arXiv preprint arXiv:2503.10677},
  year={2025}
}

@inproceedings{43,
  title={Retrieval-augmented layout transformer for content-aware layout generation},
  author={Horita, Daichi and Inoue, Naoto and Kikuchi, Kotaro and Yamaguchi, Kota and Aizawa, Kiyoharu},
  booktitle={Proceedings of the IEEE/CVF Conference on Computer Vision and Pattern Recognition (CVPR)},
  pages={67--76},
  year={2024}
}

@article{44,
  title={Chain-of-thought prompting elicits reasoning in large language models},
  author={Wei, Jason and Wang, Xuezhi and Schuurmans, Dale and Bosma, Maarten and Xia, Fei and Chi, Ed and Le, Quoc V and Zhou, Denny and others},
  journal={Advances in Neural Information Processing Systems (NeurIPS)},
  volume={35},
  pages={24824--24837},
  year={2022}
}

@article{45,
  title={Towards reasoning era: A survey of long chain-of-thought for reasoning large language models},
  author={Chen, Qiguang and Qin, Libo and Liu, Jinhao and Peng, Dengyun and Guan, Jiannan and Wang, Peng and Hu, Mengkang and Zhou, Yuhang and Gao, Te and Che, Wangxiang},
  journal={arXiv preprint arXiv:2503.09567},
  year={2025}
}

@article{46,
  title={Deconstructing Long Chain-of-Thought: A Structured Reasoning Optimization Framework for Long {CoT} Distillation},
  author={Luo, Yijia and Song, Yulin and Zhang, Xingyao and Liu, Jiaheng and Wang, Weixun and Chen, GengRu and Su, Wenbo and Zheng, Bo},
  journal={arXiv preprint arXiv:2503.16385},
  year={2025}
}

@article{47,
  title={Composition-aware graphic layout {GAN} for visual-textual presentation designs},
  author={Zhou, Min and Xu, Chenchen and Ma, Ye and Ge, Tiezheng and Jiang, Yuning and Xu, Weiwei},
  journal={arXiv preprint arXiv:2205.00303},
  year={2022}
}

@article{49,
  title={Positionic: Unified position and identity consistency for image customization},
  author={Hu, Junjie and Han, Tianyang and Ma, Kai and Gao, Jialin and Yang, Song and He, Xianhua and Luo, Junfeng and Wei, Xiaoming and Zhang, Wenqiang},
  journal={arXiv preprint arXiv:2507.13861},
  year={2025}
}

@article{50,
  title={Beyond Words and Pixels: A Benchmark for Implicit World Knowledge Reasoning in Generative Models},
  author={Han, Tianyang and Su, Junhao and Hu, Junjie and Yang, Peizhen and Shi, Hengyu and Luo, Junfeng and Gao, Jialin},
  journal={arXiv preprint arXiv:2511.18271},
  year={2025}
}

\appendix
\section{Prompt Examples of LayoutCoT}

\subsection{Content Aware Layout Generation}

\paragraph{Stage 1: Initial Adjustment}

\begin{promptbox}{System Prompt}
\textbf{Role.} You are an advanced layout design assistant, LayoutGPT. We will refine the layout in 3 separate steps. Each step is isolated, and you only have the references provided in the current context.

\textbf{Goal.} Adjust element positions to achieve reasonable alignment without overlap.

\textbf{Procedure.}
\begin{enumerate}
    \item Adjust \emph{only} \texttt{text} and \texttt{logo} elements (modify x/y position, size, and font-size).
    \item Keep \texttt{object} and \texttt{underlay} locked (do not move or alter them).
    \item Avoid overlap between \texttt{text}, \texttt{logo}, and \texttt{object} --- even 1px overlap is disallowed.
    \item Achieve reasonable alignment: for each logical group, \texttt{logo} should ideally appear above the corresponding \texttt{text}.
    \item Return only updated HTML content with no extra explanation.
\end{enumerate}

\textbf{Note.} You may learn spatial relationships and layout logic from the reference layouts, but you must not directly copy element coordinates.
\end{promptbox}

\begin{promptbox}{User Prompt}
Below are {{$LEN_{TOPK}$}} reference layouts:
\begin{lstlisting}[style=promptlisting]
{{TOPK_HTML_STR}}
\end{lstlisting}

Here is the initial layout needing refinement:
\begin{lstlisting}[style=promptlisting]
{{CURRENT_HTML}}
\end{lstlisting}
\end{promptbox}

\paragraph{Stage 2: Underlay Refinement}

\begin{promptbox}{System Prompt}
\textbf{Role.} You are LayoutGPT. Continue refining the layout.

\textbf{Goal.} Optimize the background elements based on the locked foreground.

\textbf{Procedure.}
\begin{enumerate}
    \item Keep \texttt{object}, \texttt{text}, and \texttt{logo} elements exactly as they were after Stage 1 (locked).
    \item Adjust \emph{only} \texttt{underlay} elements (background rectangles).
    \item Avoid unintended overlap: \texttt{underlay} must not intersect with unrelated content.
    \item Improve layout metrics:
    \begin{itemize}
        \item \texttt{metrics\_und\_l}: If \texttt{underlay} > target, maximize IoU with associated \texttt{text}/\texttt{logo}.
        \item \texttt{metrics\_und\_s}: If \texttt{underlay} $\le$ target, ensure it fully contains the corresponding \texttt{text}/\texttt{logo}.
    \end{itemize}
    \item Return only updated HTML content.
\end{enumerate}
\end{promptbox}

\begin{promptbox}{User Prompt}
Below are the {{$LEN_{TOPK}$}} reference layouts:
\begin{lstlisting}[style=promptlisting]
{{TOPK_HTML_STR}}
\end{lstlisting}

Here is the layout after Stage 1:
\begin{lstlisting}[style=promptlisting]
{{STAGE_1_HTML}}
\end{lstlisting}

Please do a final check and minimal fixes as needed. Return only the final HTML.
\end{promptbox}

\paragraph{Stage 3: Final Check}

\begin{promptbox}{System Prompt}
\textbf{Role.} You are LayoutGPT. Perform the final check.

\textbf{Goal.} Ensure zero unintended overlap and perfect alignment.

\textbf{Procedure.}
\begin{enumerate}
    \item Check for unintended overlap across all elements (\texttt{text}, \texttt{logo}, \texttt{underlay}, \texttt{object}). Ensure \texttt{underlay} only overlaps its own group.
    \item Ensure alignment is preserved (e.g., \texttt{logo} above \texttt{text}).
    \item Perform minimal fixes for small issues (e.g., slight 1-2px overlap, container issues).
    \item Return only the final HTML result.
\end{enumerate}
\end{promptbox}

\begin{promptbox}{User Prompt}
Below are the {{$LEN_{TOPK}$}} reference layouts:
\begin{lstlisting}[style=promptlisting]
{{TOPK_HTML_STR}}
\end{lstlisting}

Here is the layout after Stage 2:
\begin{lstlisting}[style=promptlisting]
{{STAGE_2_HTML}}
\end{lstlisting}

Please do a final check and minimal fixes as needed. Return only the final HTML.
\end{promptbox}

\subsection{Constraint-Explicit Layout Generation (Gen-T)}
We provide the prompt for the Gen-T task on the PubLayNet dataset as a representative example.

\paragraph{Stage 1: Main Element Placement}

\begin{promptbox}{System Prompt}
\textbf{Context.} We have 5 element types: 1) \texttt{text}, 2) \texttt{title}, 3) \texttt{list}, 4) \texttt{table}, 5) \texttt{figure}.

\textbf{Goal.} Initial placement of core text elements.

\textbf{Procedure.}
\begin{enumerate}
    \item Move/resize \emph{only} \texttt{title} (2) and \texttt{text} (1).
    \item Keep \texttt{list} (3), \texttt{table} (4), and \texttt{figure} (5) locked.
    \item Minimize overlap between \texttt{title} \& \texttt{text} and place them with good alignment.
    \item Keep FID \& maxIoU in mind.
    \item Return only updated HTML.
\end{enumerate}
\end{promptbox}

\begin{promptbox}{User Prompt}
Here are {{$LEN_{TOPK}$}} reference layouts:
\begin{lstlisting}[style=promptlisting]
{{TOPK_HTML_STR}}
\end{lstlisting}

Initial layout needing Stage 1 refinement:
\begin{lstlisting}[style=promptlisting]
{{CURRENT_HTML}}
\end{lstlisting}

Please adjust only \texttt{title}(2) \& \texttt{text}(1) to reduce overlap and improve alignment.
\end{promptbox}

\paragraph{Stage 2: Complex Element Placement}

\begin{promptbox}{System Prompt}
\textbf{Goal.} Place complex elements while respecting Stage 1.

\textbf{Procedure.}
\begin{enumerate}
    \item Place/resize \texttt{list} (3), \texttt{table} (4), and \texttt{figure} (5).
    \item Keep \texttt{title} (2) \& \texttt{text} (1) from Stage 1 locked if possible.
    \item Further reduce overlap among all 5 elements.
    \item Refine alignment (e.g., align edges of tables with text).
    \item Begin improving FID and maxIoU.
    \item Return only updated HTML.
\end{enumerate}
\end{promptbox}

\begin{promptbox}{User Prompt}
{{$LEN_{TOPK}$}} reference layouts:
\begin{lstlisting}[style=promptlisting]
{{TOPK_HTML_STR}}
\end{lstlisting}

Layout after Stage 1:
\begin{lstlisting}[style=promptlisting]
{{STAGE_1_HTML}}
\end{lstlisting}

Please refine \texttt{list}(3), \texttt{table}(4), \texttt{figure}(5) now. Return updated HTML.
\end{promptbox}

\paragraph{Stage 3: Global Optimization}

\begin{promptbox}{System Prompt}
\textbf{Goal.} Final thorough check for metrics optimization.

\textbf{Objectives.}
\begin{itemize}
    \item \textbf{Alignment:} Lower score implies better alignment.
    \item \textbf{Overlap:} Lower score implies minimal collisions.
    \item \textbf{FID:} Reduce unnatural layout styles.
    \item \textbf{maxIoU:} Match reference bounding arrangements.
\end{itemize}

\textbf{Procedure.}
\begin{enumerate}
    \item Enumerate each pair for overlap.
    \item Check alignment lines or columns.
    \item Evaluate if FID can be lowered by slight repositioning.
    \item Shift/resize elements to improve IoU with references.
    \item Provide brief chain-of-thought reasoning, then output final HTML.
\end{enumerate}
\end{promptbox}

\begin{promptbox}{User Prompt}
{{$LEN_{TOPK}$}} reference layouts:
\begin{lstlisting}[style=promptlisting]
{{TOPK_HTML_STR}}
\end{lstlisting}

Layout after Stage 2:
\begin{lstlisting}[style=promptlisting]
{{STAGE_2_HTML}}
\end{lstlisting}

Perform your step-by-step reasoning to finalize alignment, overlap, FID, maxIoU. Return final HTML.
\end{promptbox}

\subsection{Text-to-Layout Generation}

\paragraph{Stage 1: Parsing and Rough Layout}

\begin{promptbox}{System Prompt}
\textbf{Role.} You are LayoutGPT.

\textbf{Goal.} Parse text and establish initial structure.

\textbf{Procedure.}
\begin{enumerate}
    \item Parse \texttt{text} to identify required elements; add missing or remove extraneous ones.
    \item Attempt strong overlap reduction from the start (moves or resizes).
    \item Perform basic alignment ($\pm$10px).
    \item Partially match reference bounding arrangement for mIoU if feasible.
    \item Return updated HTML only.
\end{enumerate}
\end{promptbox}

\begin{promptbox}{User Prompt}
References (topk) total {{$LEN_{TOPK}$}}:
\begin{lstlisting}[style=promptlisting]
{{REFERENCES_STR}}
\end{lstlisting}

Current layout:
\begin{lstlisting}[style=promptlisting]
{{CURRENT_HTML}}
\end{lstlisting}

Text describing needed elements:
\begin{lstlisting}[style=promptlisting]
{{TEXT_DESCRIPTION}}
\end{lstlisting}

Please add/remove elements per text, reduce overlap, and do basic alignment. Return updated HTML.
\end{promptbox}

\paragraph{Stage 2: Overlap Resolution}

\begin{promptbox}{System Prompt}
\textbf{Goal.} Aggressive overlap reduction.

\textbf{Procedure.}
\begin{enumerate}
    \item Reduce Overlap to near zero (Top Priority).
    \item Refine alignment lines ($\pm$5px).
    \item Match bounding arrangement from references for higher mIoU.
    \item Add or remove minor elements if beneficial.
    \item Return updated HTML only.
\end{enumerate}
\end{promptbox}

\begin{promptbox}{User Prompt}
References:
\begin{lstlisting}[style=promptlisting]
{{REFERENCES_STR}}
\end{lstlisting}

Layout after Stage 1:
\begin{lstlisting}[style=promptlisting]
{{STAGE_1_HTML}}
\end{lstlisting}

Focus on Overlap $\to$ 0, align row/col $\pm$5px. Return updated HTML.
\end{promptbox}

\paragraph{Stage 3: Semantic-Agnostic Refinement}

\begin{promptbox}{System Prompt}
\textbf{Goal.} Final geometric polish.

\textbf{Procedure.}
\begin{enumerate}
    \item Temporarily ignore text meaning. Force Overlap = 0 among main elements.
    \item Fix leftover collisions by moves/resizes.
    \item Push bounding arrangement for better mIoU.
    \item Fix FID style if it looks awkward.
    \item Return updated HTML only.
\end{enumerate}
\end{promptbox}

\begin{promptbox}{User Prompt}
References:
\begin{lstlisting}[style=promptlisting]
{{REFERENCES_STR}}
\end{lstlisting}

Layout after Stage 2:
\begin{lstlisting}[style=promptlisting]
{{STAGE_2_HTML}}
\end{lstlisting}

Ignore text semantics, finalize Overlap=0, fix FID. Return updated HTML.
\end{promptbox}
\end{document}